\documentclass{article}

\usepackage{microtype}
\usepackage{graphicx}
\usepackage{subfigure}
\usepackage{booktabs} % for professional tables

\usepackage{hyperref}

\usepackage[noend]{algorithm2e}

\usepackage[accepted]{icml2023}

\usepackage{amsmath}
\usepackage{amssymb}
\usepackage{mathtools}
\usepackage{amsthm}

\usepackage{wrapfig, multirow}
\usepackage{xcolor}

\usepackage[capitalize,noabbrev]{cleveref}

\theoremstyle{plain}
\newtheorem{theorem}{Theorem}[section]

\theoremstyle{definition}
\newtheorem{definition}[theorem]{Definition}

\theoremstyle{remark}

\usepackage[textsize=tiny]{todonotes}

\icmltitlerunning{Towards customizable reinforcement learning agents: Enabling preference specification through online vocabulary expansion}

\begin{document}
\newcommand{\edit}[1]{\textcolor{red}{#1}}
\twocolumn[
\icmltitle{Towards customizable reinforcement learning agents: Enabling preference specification through online vocabulary expansion}



\icmlsetsymbol{equal}{*}

\begin{icmlauthorlist}
\icmlauthor{Utkarsh Soni}{yyy}
\icmlauthor{Nupur Thakur}{yyy}
\icmlauthor{Sarath Sreedharan}{sch}
\icmlauthor{Lin Guan }{yyy}
\icmlauthor{Mudit Verma}{yyy}
\icmlauthor{Matthew Marquez}{yyy}
\icmlauthor{Subbarao Kambhampati}{yyy}
% \icmlauthor{Arizona State University}{yyy}
% \icmlauthor{Colorado State University}{sch}
\end{icmlauthorlist}

% \author[1]{}
% \author[2]{}
% \author[1]{}
% \author[1]{}
% \author[1]{}
% \author[1]{ }
% \affil[1]{}
% \affil[2]{}

\icmlaffiliation{yyy}{Arizona State University}
\icmlaffiliation{sch}{Colorado State University}

\icmlcorrespondingauthor{Utkarsh Soni}{usoni1@asu.edu}

\icmlkeywords{Machine Learning, ICML}

\vskip 0.3in
]



\printAffiliationsAndNotice{}  % leave blank if no need to mention equal contribution

\begin{abstract}
There is a growing interest in developing automated agents that can work alongside humans. In addition to completing the assigned task, such an agent will undoubtedly be expected to behave in a manner that is preferred by the human. This requires the human to communicate their preferences to the agent. To achieve this, the current approaches either require the users to specify the reward function or the preference is interactively learned from queries that ask the user to compare behavior. The former approach can be challenging if the internal representation used by the agent is inscrutable to the human while the latter is unnecessarily cumbersome for the user if their preference can be specified more easily in symbolic terms. In this work, we propose \textit{PRESCA} (PREference Specification through Concept Acquisition), a system that allows users to specify their preferences in terms of concepts that they understand. \textit{PRESCA} maintains a set of such concepts in a shared vocabulary. If the relevant concept is not in the shared vocabulary, then it is learned. To make learning a new concept more feedback efficient, \textit{PRESCA} leverages causal associations between the target concept and concepts that are already known. In addition, we use a novel data augmentation approach to further reduce required feedback. We evaluate \textit{PRESCA} by using it on a Minecraft environment and show that it can effectively align the agent with the user's preference. 
\end{abstract}

\section{Introduction}

% Things to add  - specific feedback, system that keeps collecting data and we can retain concepts from previous interactions, causal model
With recent successes in AI, there is a great interest in deploying autonomous agents into our day to day lives. In order to cohabit successfully with humans, it is highly important that the AI agent behaves in a way that is aligned with the human preferences. Ideally, we want a system that will enable every day users to specify their preferences over AI system behavior. In the reinforcement learning (RL) literature, the current go to approach for specifying behavioral preferences is through preference-based reinforcement learning techniques (\cite{christiano2017deep}; \cite{lee2021pebble}) that try to learn the human's preference interactively through trajectory comparisons. These techniques are useful for tacit knowledge tasks. However, it would be highly inefficient to use these techniques in scenarios where the preference can simply be specified in symbolic terms. Another way for specifying behavioral preferences is through modifying rewards; but it can be fairly non-intuitive for a lay user to come up with a reward structure that leads to the preferred behavior \cite{hadfield2017inverse}. In addition, specifying rewards becomes more challenging when the system is operating over an inscrutable high-dimensional state representation (like images). An alternate is to allow humans to specify preferences in symbolic terms. 
These symbolic concepts can be propositional state variables that the user understands. Thus, a more suitable framework that lets user specify their preferences would consist of a symbolic interface made of such concepts that enables communication with the user while the agent uses some inscrutable internal representation for the task.  

There already exists a line of works (\cite{lyu2019sdrl}; \cite{illanes2020symbolic}; \cite{icarte2022reward}) that let the user specify task-related information to an RL agent in symbolic terms. Unfortunately, these works assume that all concepts relevant to the task information are already known, i.e., the grounding of each of these concepts is available. However, since each user's preference can be unique, their preference could involve concepts that are not present in the agent's vocabulary. In this work, we propose an AI system named \textit{PRESCA} (PREference Specification through Concept Acquisition) that maintains a symbolic interface made of concepts that the user can use to specify their preferences to the agent. If the concept that is relevant to the user's preference is missing from the interface, then \textit{PRESCA} will try to learn this concept online. Subsequently, the concept is also added to the interface to support preference specifications by future users. Thus, the cost of learning the concept gets amortized when future users make use of the concept. Once the preference has been specified, \textit{PRESCA} uses it to train the agent to align with the user's preferences. The focus of this work, will be to propose a method that allows  us to learn  human concepts effectively. A simple way for the system to learn a concept is to learn a grounding from the concept to system states. One could learn these groundings from a set of positive and negative examples of the concept. Obtaining these examples is a challenging problem as there is no clear way for the user to generate these examples. One way to obtain these examples is for the system to present the user with states and ask them whether the concept is present in each state. A naive way to generate these queries could be to randomly sample states from the environment. However, if the positive examples are sparse in the state space, then this strategy would lead to a possibly large number of queries. 
% \begin{wrapfigure}{R}{0.6\textwidth}
%\centering
\begin{figure}
\includegraphics[scale=0.2]{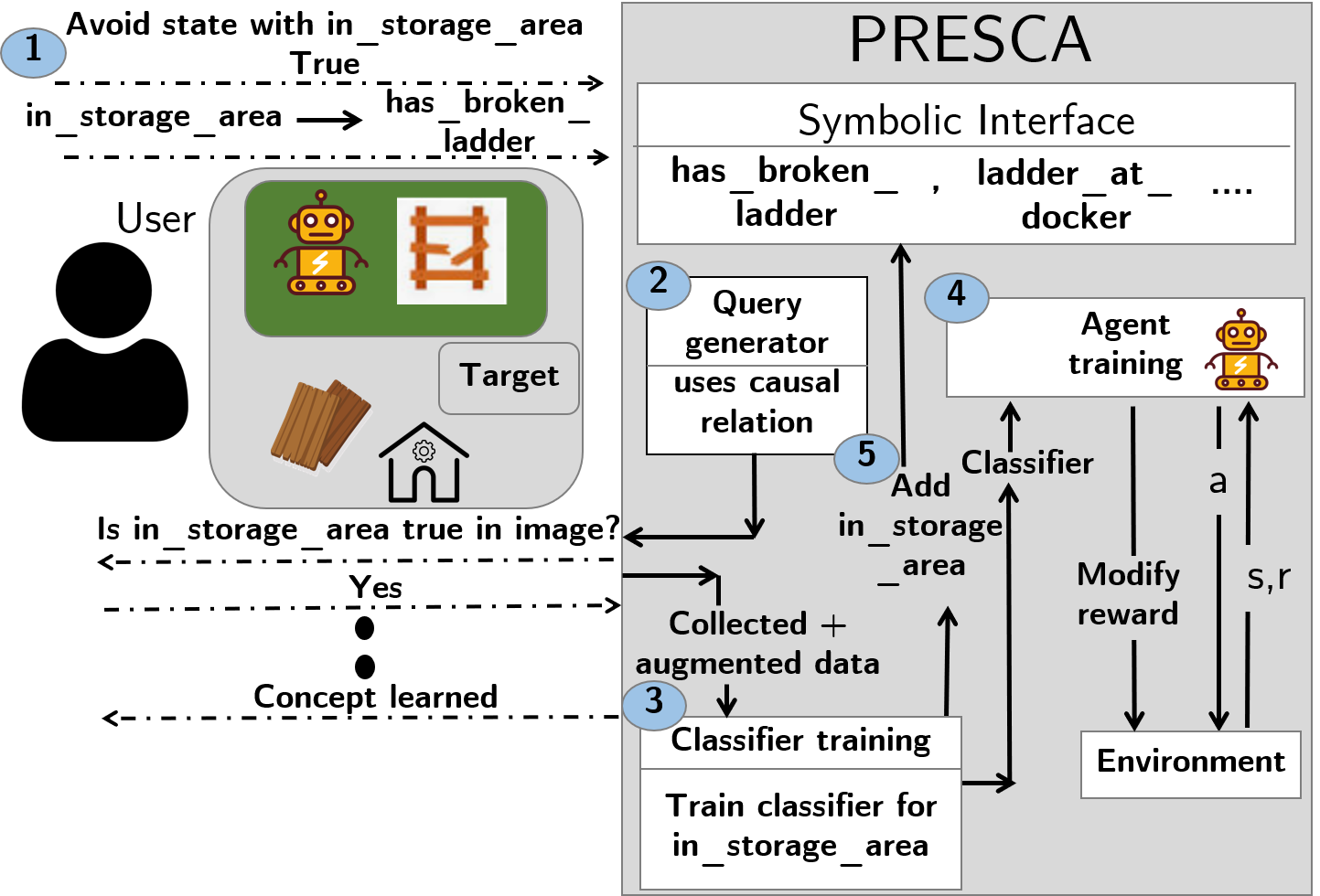}
\caption{Overview of \textit{PRESCA}. (1) The user specifies their preference in terms of some symbolic concept. If the concept is not present in the symbolic interface, then the user provides its causal relationship to some known concept. (2) \textit{PRESCA} then generates likely positive examples and negative examples of the concept and queries their label to the user. (3) After getting the labels, and data augmentation \textit{PRESCA} learns a classifier for the target concept and (4) incorporates user's preference in agent's training. (5) Finally, the concept is added to the interface.}
\label{fig:1}

\end{figure}

Our objective is to make the data collection process more feedback efficient by automatically gathering likely positive and negative examples of the concept and then querying the user for their labels. To this end, we leverage the causal association the target concept has with the concepts that already exist in the symbolic interface. This causal knowledge can be automatically obtained if a symbolic PDDL-like model is available (\cite{geffner2013concise}; \cite{helmert2004planning}). If such a domain model is not present, then we expect the user of the system to have some understanding about the task dynamics and provide the causal relationship between the target concept and some known concept. In addition to leveraging the causal relationship to reduce human labeling effort, \textit{PRESCA} also uses a novel data augmentation strategy to reduce the amount of labels needed from the user. In figure \ref{fig:1}, we provides the overall flow of the user's interaction with the \textit{PRESCA} system.  In the following section, we first introduce the planning domain that we use to illustrate and evaluate our approach. This is followed by a formal description of the environment model and the symbolic interface used by our AI system. We then provide a methodology that efficiently learns a new concept and then uses the learned concept to guide the agent's training. In the evaluation section (section \ref{evaluation}), we show the performance of our approach on a Minecraft domain given increasingly complex causal relations. We follow that up with a discussion that compares our approach to existing AI approaches that can also be potentially used to incorporate user's preference. Finally, we discuss the possible improvements in section \ref{planned_improvements}. 

\section{Illustrative example}

% \begin{wrapfigure}{R}{0.6\textwidth}
%\centering
% \begin{figure}
%  \includegraphics[scale=0.15]{Images/domain_image_2.png}
% \caption{(a) Instance of the Minecraft environment with two possible plans marked with arrows. The user prefers that the agent avoid going into the storage area (indicated by the red arrows) (b) the causal model of the Minecraft environment. }
% \label{fig:2}
% \end{figure}

% \setlength{\columnsep}{7pt}%

\begin{wrapfigure}{r}{0.30\textwidth}
	\begin{center}
		\includegraphics[width=0.30\textwidth]{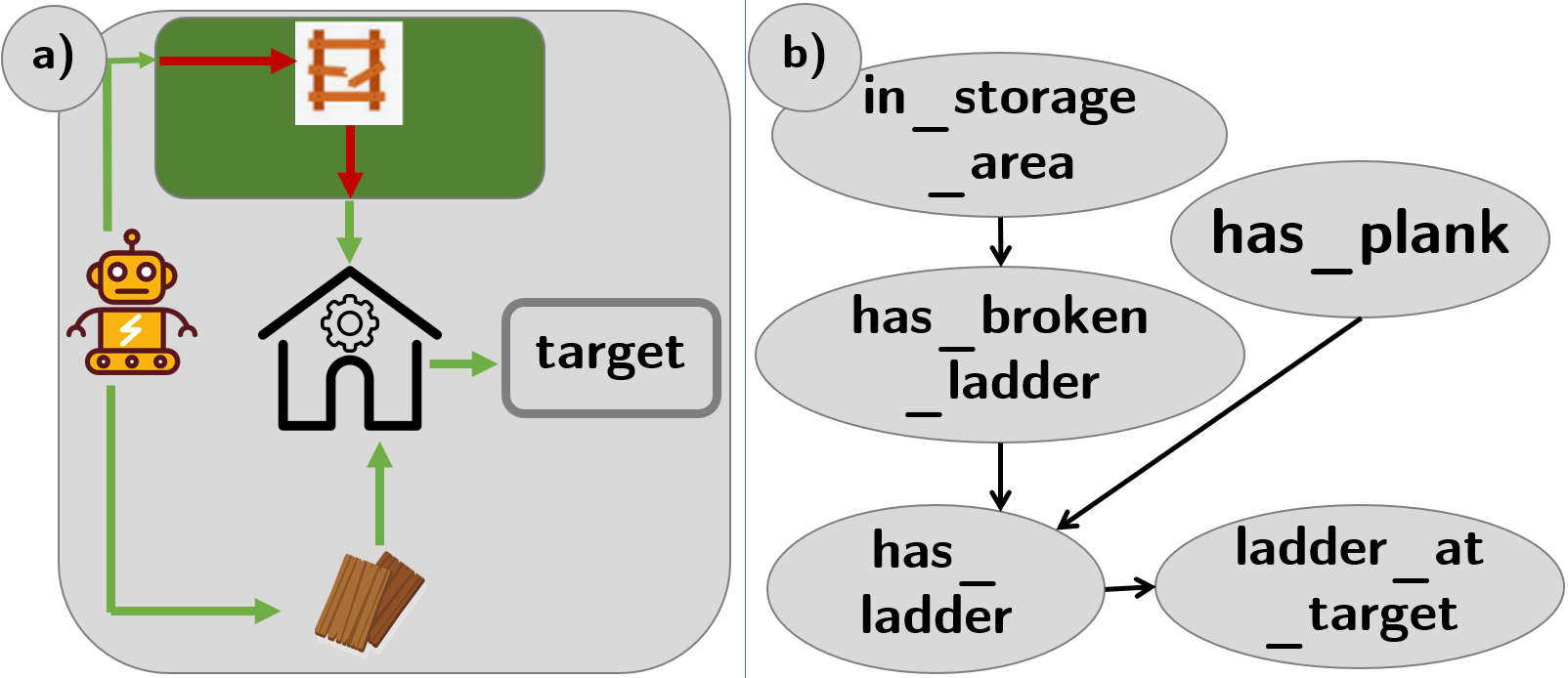}
        
        \caption{(a) Instance of the Minecraft environment with two possible plans marked with arrows. The user prefers that the agent avoid going into the storage area (indicated by the red arrows) (b) the causal model of the Minecraft environment. }
        \label{fig:2}
	\end{center}
    
\end{wrapfigure}

We will use a version of the 2-D Minecraft environment \cite{andreas2017modular} (figure \ref{fig:2} (a)) to illustrate the ideas of the paper and evaluate our proposed technique. In this domain, the goal is to drop a ladder at the target. The set of actions the agent can take include, turning left or right, moving forward, picking an object, a crafting action, a no-op action, and an action that drops the ladder when the agent is at the target. To accomplish the goal, the agent needs to first obtain a ladder. There are two ways to achieve this. One way is for the agent to pick up a plank; and  then use the crafting station to craft a ladder. Another way is to first move into the storage area (green region in figure \ref{fig:2}(a)), then pick up a broken ladder and then use the crafting station to repair the ladder. Once the agent has the ladder, it can move to the target and drop it. We show the two possible plans in figure \ref{fig:2}(a).  There is also a human observer who wants the agent to solve the task in the way they prefer. The observer wants the agent to avoid going inside the storage area. Since one way of solving the task involves moving into the storage area, the human must communicate this preference to the agent. In this scenario, the agent is operating on some state encoding that the human cannot interpret. Thus, they cannot specify their preference directly in terms of state features. However, if the human is capable of communicating with the AI system in symbolic terms, they could simply ask the system to avoid any state where the fact \textit{in\_storage\_area} might be true. In this case, \textit{in\_storage\_area} is a propositional fact that is \textit{true} in every state where the agent is inside the storage area. To support such a symbolic specification, the AI system should be able to correctly \textit{ground}  and thereby interpret the potential concepts that the user wants to use. In this case, \textit{PRESCA} learns a classifier that can predict whether in a given state the agent is in the storage area. Learning this classifier requires the user to provide positive and negative examples of the concept \textit{in\_storage\_area}. In this work, we make this data collection process more feedback efficient. To do so, we leverage the precedence relation \textit{in\_storage\_area} has with other concepts that are already known. Figure \ref{fig:2} (b), illustrates the causal relationship between various concepts in the domain. For e.g., the concept \textit{in\_storage\_area} precedes \textit{has\_broken\_ladder}. This causal link implies that the agent must be inside the storage area to collect the broken ladder. Note that there can also be concepts with multiple possible causes like  the \textit{has\_ladder} concept.  Now, if any of the descendents of the concept \textit{in\_storage\_area} is already known, then our proposed method tries to use this causal knowledge to gather likely positive examples of the concept \textit{in\_storage\_area}.

\section{Problem Setting and assumptions}
We consider an RL problem in which an agent interacts with an unknown environment (\cite{sutton2018reinforcement}). The environment is modeled as a Markov Decision Process (MDP). An MDP $\mathcal{M}$ can be formally defined as tuple $M = \langle S, A, T, R, \gamma, S_o \rangle$, where: $S$ is the set of states in the environment, $A$ is the set of actions that the agent can take, $T$ is the transition function where $T(s, a, s')$ gives the probability that agent will be in state $s'$ after taking action $a$ in state $s$, $R$ is the reward function where $R(s, a, s')$ gives the reward obtained for the transition $\langle s, a, s' \rangle$, $\gamma$ is the discounting factor, and $S_o$ is the set of all possible initial states. A policy $\pi (a | s)$ gives the probability that the agent will take action $a \in A$ while in the state $s \in S$. The value of a state $s$  given a policy $\pi$ , $V_{\pi}(s)$ is the expected cumulative discounted future reward that the agent obtains when following $\pi$ from the state $s$. For an MDP, $\mathcal{M}$ , the optimal policy is the policy that maximizes the value for every state. Our setting additionally considers the agent to be goal-directed. This means that there is a set of goal states $\mathbb{G}$ and the agent tries to reach one of them. We assume that the reward function $R$ is set up in a way that any optimal policy must reach one of the goal states with probability $=1$. 

In this work, we are interested in a human-AI interaction setting, where the agent will interact with multiple users over its lifetime. In each interaction, there will be a human-in-the-loop, who wants the agent to achieve the goal subject to their preferences.  Thus, we develop the system, \textit{PRESCA}, that would allow any user to communicate their preference to the agent. Now, the state representation used in the model $\mathcal{M}$ may be inscrutable to a user i.e. the user cannot directly use it to specify their preference (e.g., an image based state representation). Therefore, we consider the presence of a symbolic interface which is a set $F_S$ of propositional state variables or concepts that the any user would understand. In any given state $s$, each concept $C_{\mathbb{Z}} \in F_S$ may be \textit{true} or \textit{false}. In \textit{PRESCA}, we currently support specifying preferences that involve the agent avoiding states where some specified concept is \textit{true}. \textit{PRESCA} can be easily extended to support more complex preferences (section \ref{training}). 

\textit{PRESCA} starts with some partial vocabulary i.e. some concepts are already present in $F_S$ and the accurate grounding for these initial concept is available. This isn't unlike other \textit{neuro-symbolic} approaches that assume known concepts (\cite{lyu2019sdrl}, \cite{illanes2020symbolic}, \cite{icarte2022reward}). However, in a specific interaction with a user, if their preference cannot be specified in terms of any existing concept in $F_S$, then \textit{PRESCA} supports learning the relevant novel concept, $C_{\mathbb{T}}$. By learning a concept $C_{\mathbb{T}}$, we mean learning some grounding in the state representation used by the agent. In our case, this grounding takes the form of a binary classifier that takes as input a state image $s$ and outputs \textit{true} if the concept $C_{\mathbb{T}}$ is \textit{true} in $s$ and \textit{false} otherwise. From now on we will use the notation $C_{\mathbb{Z}}$ to refer to both the concept as well as its grounding. Now to learn the classifier for the target concept $C_{\mathbb{T}}$, the system must collect positive and negative examples of the concept from the user. For this, our system presents the user with state queries where they must choose whether the concept is present or absent in the state. Once the classifier $C_{\mathbb{T}}$ has been learned, it is used to train an agent's policy that aligns with the user's preference. Also, the classifier $ C_{\mathbb{T}}$ is added to the set $F_S$ to support preference specifications by future users. We now precisely define the objective of the \textit{PRESCA} system for a single interaction with a user when some new concept needs to be learned. Given the environment $\mathcal{M}$ , a user, a symbolic interface $F_S$ , and a target concept $C_{\mathbb{T}}$, the system must: (a) minimize the number of queries made to the user to learn $C_{\mathbb{T}}$, and (b) use $C_{\mathbb{T}}$ to train an agent's policy that achieves the goal while aligning with the user's preference. 

\subsection{Causal model semantics}
The \textit{PRESCA} system uses the causal relationship between the target concept $C_{\mathbb{T}}$ and some already known concept $C_{\mathbb{K}} \in F_S$ to gather candidate states that most likely contain $C_{\mathbb{T}}$. The causal relationship is simply a partial causal model of the domain. Intuitively, the causal model for a domain is a directed graph with nodes representing concepts and edges representing some causal association. Figure \ref{fig:2}(b) shows an example causal model for the Minecraft domain. In the causal model, the connections between some concept $C_\mathbb{Z}$ and all its parents concepts reflect an abstract relationship the parent concepts have with $C_\mathbb{Z}$ in the dynamics of the domain. Informally, the relation $C_{\mathbb{T}} \rightarrow C_{\mathbb{K}}$, where $C_{\mathbb{T}}$ is the only parent of $C_{\mathbb{K}}$, dictates that if in a transition $\langle s, a, s'\rangle$ the concept $C_{\mathbb{K}}$ is \textit{true} in $s'$ while it was \textit{false} in $s$, then $C_{\mathbb{T}}$ must have been true in the state $s$ i.e. $C_{\mathbb{T}}$ must be \textit{true} in the current state for $C_{\mathbb{K}}$ to be \textit{true} in the next state. One can understand such causal relations by relating them to preconditions and effects from symbolic models (like PDDL \cite{geffner2013concise}). In this case, the relation $C_{\mathbb{T}} \rightarrow C_{\mathbb{K}}$, may be due to some abstract symbolic action whose precondition is $C_{\mathbb{T}}$ and one of the effects are $C_{\mathbb{K}}$. 

The causal model itself is represented using a syntax that is similar to that of structured causal models \cite{glymour2016causal}.  Formally, we define the causal model $\mathcal{M}_{S}$ as a tuple $\langle V_{S}, E_{S}, f_{S} \rangle$ where $V_S$ and $E_S$ correspond to nodes and edges of a directed graph where each node $v \in V_S$ corresponds to a concept; and $f_{S}$ is the set of structural equations specifying an abstract relationship between a child concept and its parents concepts. Each equation in $f_S$, that relates the parent concepts $C_{\mathbb{X}_1}, C_{\mathbb{X}_2} \dots C_{\mathbb{X}_n}$ to their common child concept $C_{\mathbb{Y}}$, is of the form $C_{\mathbb{Y}} = \mathcal{V}(C_{\mathbb{X}_1}, C_{\mathbb{X}_2} \dots C_{\mathbb{X}_n})$ where $\mathcal{V}$ is a boolean function. We limit $\mathcal{V}$ to be disjunctions of positive literals but the extension to more complex formulae is straightforward. We denote the boolean function $\mathcal{V}$ in the structural formulae of a concept $C_{\mathbb{Y}}$ as $\mathcal{V}_{C_{\mathbb{Y}}}$.  Now, for any concept $C_{\mathbb{Z}}$, let $C_{\mathbb{Z}}(s)$ denote the predicate that indicates whether or not $C_{\mathbb{Z}}$ is \textit{true} in the state $s$. Given this notation, we now define the grounding of each structural equation in transitions within the domain.

\begin{definition}
\label{def1}
Given a concept $C_{\mathbb{Y}} \in {V}_{S}$, its structural equation $C_{\mathbb{Y}} = \mathcal{V}(C_{\mathbb{X}_1}, C_{\mathbb{X}_2} \dots C_{\mathbb{X}_n})$ and a transition $\tau = \langle s, a, s'\rangle$ in $\mathcal{M}$,  if $C_{\mathbb{Y}}(s')$ is \textit{true} and $C_{\mathbb{Y}}(s)$ is \textit{false}, then the grounding of the structural equation is defined as the following equation that holds for $\tau$: $C_{\mathbb{Y}}(s') = \mathcal{V}(C_{\mathbb{X}_1}(s), C_{\mathbb{X}_2}(s) \dots C_{\mathbb{X}_n}(s))$. 
\end{definition}

For instance, in the Minecraft domain (figure \ref{fig:2}), the structural equation for \textit{has\_ladder} is of the form $C_{has\_ladder} = C_{has\_broken\_ladder} \lor C_{has\_planks}$ . Thus, if there is a transition where the agent picks up a ladder, then from definition \ref{def1}, we know that the agent either had the broken ladder; or it had planks in the original state.

\subsection{Concept locality and rarity}
\label{assumptions}
\textit{PRESCA} also leverages simple assumptions about concepts for data collection. We make a \textit{concept locality} assumption: if a concept $C_{\mathbb{Z}}$ is true in some state $s$, then it would most likely be true in a small neighborhood around $s$. More specifically, if we take the set of all states $S'$ where each state $s' \in S'$ can be reached from $s$ within some small number of actions, then the states in $S'$ will likely have the concept $C_{\mathbb{Z}}$ \textit{true} in them. The intuition behind this assumption is that the agent would usually have to take some specific sequence of actions to make the concept be no longer  \textit{true}. Another assumption we make is about  \textit{concept rarity}. Specifically, for any concept $C_{\mathbb{Z}}$, the proportion of states in which the concept $C_{\mathbb{Z}}$ is \textit{true} will be low i.e. $|S_{C_{\mathbb{Z}}}| / |S| \ll 1$ where $S_{C_{\mathbb{Z}}}$ is the set of states with $C_{\mathbb{Z}}$ \textit{true}. 
\section{Learning concept classifier and preferred policy}
% Our focus in this work is to make the data collection process for learning the target concept $C_{\mathbb{T}}$ more efficient. For the agent training, we rely on straightforward reward shaping and option learning techniques. We start by describing the semantics of a causal model and stating some simple assumptions about concepts that \textit{PRESCA} uses to gather likely positive and negative examples of $C_{\mathbb{T}}$. We then describe both the data collection, and agent training process in detail.  

We now begin to describe our approach to both learning the target concept classifier $C_{\mathbb{T}}$, as well as, how the preference is incorporated in the policy. Our data collection approach involves gathering likely positive and negative examples of the concept $C_{\mathbb{T}}$. As mentioned earlier, this would reduce the total queries made to the user when compared to an approach where states are sampled randomly and then queried to the user.  For collecting likely positive examples of $C_{\mathbb{T}}$, we use the causal relation between the target $C_{\mathbb{T}}$ and some known concept $C_{\mathbb{K}} \in F_S$. The required causal relationship can either be given by the user or it can be derived from a symbolic domain model using \cite{helmert2004planning} if the domain model is available (as is the case in multiple \textit{neuro-symbolic} techniques like \cite{illanes2020symbolic}). The causal relation is a partial specification of the causal model $\mathcal{M}_S$ of the domain, denoted as $\mathcal{M}_{\mathbb{T} \rightarrow \mathbb{K}}$ where $\mathcal{M}_{\mathbb{T} \rightarrow \mathbb{K}} = \langle V_{\mathbb{T} \rightarrow \mathbb{K}}, E_{\mathbb{T} \rightarrow \mathbb{K}}, f_{\mathbb{T} \rightarrow \mathbb{K}}\rangle$ such that $\langle V_{\mathbb{T} \rightarrow \mathbb{K}}, E_{\mathbb{T} \rightarrow \mathbb{K}} \rangle$ is a subgraph of $\langle V_S, E_S \rangle$ and $f_{\mathbb{T} \rightarrow \mathbb{K}} \subseteq f_S$. The causal relation $\mathcal{M}_{\mathbb{T} \rightarrow \mathbb{K}}$ must necessarily contain a path from $C_{\mathbb{T}}$ to some concept $C_{\mathbb{K}}$  where $C_{\mathbb{K}}$ is known. We denote this path as $\mathcal{P_{\mathbb{T} \rightarrow \mathbb{K}}}$. Additionally, it must also contain the structural equation of the node $C_\mathbb{K}$ and each of the intermediate nodes on the path $\mathcal{P_{\mathbb{T} \rightarrow \mathbb{K}}}$. Our approach for getting examples of $C_{\mathbb{T}}$ is relatively straightforward if $C_{\mathbb{K}}$ is the immediate child of $C_{\mathbb{T}}$. However, the process gets significantly involved when there are multiple intermediate concepts between $C_{\mathbb{T}}$ and $C_{\mathbb{K}}$. When this is the case, then our approach also involves learning the classifiers for the intermediate concepts (section \ref{approach}). 

Given the path  $\mathcal{P_{\mathbb{T} \rightarrow \mathbb{K}}}$, we start by describing distinct components of our algorithm. We describe each component of our approach using an example concept $C_{\mathbb{X}}$ which can be any of the unknown concepts on the path $\mathcal{P_{\mathbb{T} \rightarrow \mathbb{K}}}$. The first component (section \ref{seed_algo}) of our algorithm involves using the causal chain to get likely positive examples of $C_{\mathbb{X}}$ which we refer to as the \textit{seed} examples of $C_{\mathbb{X}}$. Once \textit{seed} examples are found, the other two components (section \ref{c_star} and \ref{c_hat}) use distinct ways to expand the number of examples of $C_{\mathbb{X}}$ and subsequently train the classifier for $C_{\mathbb{X}}$. We provide our complete approach is section \ref{approach} where we see how these components interact together to learn the classifier $C_{\mathbb{T}}$.

\subsection{Collecting seed examples of $C_{\mathbb{X}}$}
\label{seed_algo}
This component involves collecting examples of $C_{\mathbb{X}}$ using the causal relationships given in $\mathcal{P_{\mathbb{T} \rightarrow \mathbb{K}}}$.  We refer to the positive examples collected using some causal relation as \textit{seed} examples and the set of all \textit{seed} examples of $C_{\mathbb{X}}$ as $Seed_{C_{\mathbb{X}}}$. Let us first look at the case when the immediate child concept of $C_{\mathbb{X}}$ in $\mathcal{P_{\mathbb{T} \rightarrow \mathbb{K}}}$ is known. Let this child concept be $C_{\mathbb{Y}}$. Note that $C_{\mathbb{Y}}$ can be known in one of two ways: (1) the concept's accurate grounding was already available to the system or (2) its classifier was learned by \textit{PRESCA}. Now we use the causal relation, $ C_{\mathbb{X}} \rightarrow C_{\mathbb{Y}}$ to get likely positive examples of  $C_{\mathbb{X}}$. To do so, we let the agent explore the environment with a policy that selects actions uniformly  until it encounters a transition $\langle s, a, s' \rangle$ with the concept $C_{\mathbb{Y}}$  \textit{false} in $s$ and \textit{true} in $s'$. Now this component can make one of two inferences: (1) $C_{\mathbb{X}}$ is \textit{true} in $s$, or (2) $C_{\mathbb{X}}$ is likely \textit{true} in $s$. There are two cases when this component makes the latter inference. The first case occurs when $C_{\mathbb{Y}}$ is a classifier that was previously learned (possibly as part of our approach $\S$ \ref{approach}). Thus our prediction of $C_{\mathbb{Y}}$ could be incorrect. The second case occurs when $C_{\mathbb{X}}$ is not the only possible cause of $ C_{\mathbb{Y}}$ i.e. $\mathcal{V}_{C_{\mathbb{Y}}}$ (the boolean function in the structural formulae of $ C_{\mathbb{Y}}$) contains multiple disjuncts. For e.g., $\mathcal{V}_{C_{\mathbb{Y}}}$ could be $C_{\mathbb{X}} \lor C_{\mathbb{X'}}$. In this case,  either $C_{\mathbb{X}}$ or $C_{\mathbb{X'}}$ is true in $s$. Here we assume that in most cases, the probability that $C_{\mathbb{X}}$ is \textit{true} in $s$ will be much higher than the probability of $C_{\mathbb{X}}$ being \textit{true} in a randomly sampled state since it is one of the causes of $ C_{\mathbb{Y}}$. Therefore, in both these cases, the component picks $s$ as being a likely positive example. It then queries the user for the label of $s$, and adds it to $Seed_{C_{\mathbb{X}}}$ if it is positive. If neither of these two cases are \textit{true} i.e. the system has an accurate grounding of $C_{\mathbb{Y}}$ and $\mathcal{V}_{C_{\mathbb{Y}}} = C_{\mathbb{X}}$, then the component infers that the $C_{\mathbb{X}}$ is \textit{true} in $s$ and adds $s$ to $Seed_{C_{\mathbb{X}}}$. This entire process is repeated until we have $\mathcal{U}$ seed examples of  $C_{\mathbb{X}}$. In case the immediate child of $C_{\mathbb{X}}$ is not known, \textit{PRESCA} first learns the concept classifier for the child concept. It then uses the learned classifier to collect $Seed_{C_{\mathbb{X}}}$(section \ref{approach}). 

\subsection{Classifier trained with informed queries, $C_{\mathbb{X}}^{*}$}
\label{c_star}
In this component, given the seed examples of concept $C_{\mathbb{X}}$, we make informed queries to expand the number of examples of $C_{\mathbb{X}}$. We leverage \textit{concept locality} to collect likely positive examples of $C_{\mathbb{X}}$. For this, we perform random walk of short length from a randomly sampled state from $Seed_{C_{\mathbb{X}}}$. The trajectory's states are then queried to the user for their label. This is repeated till $N^{+}$ positive examples have been collected. Given \textit{concept rarity} (section \ref{assumptions}), if we randomly sample a state from the state space, then it will most likely be negative. Thus, for collecting likely negative examples of $C_{\mathbb{X}}$, we simply collect randomly sampled states. More specifically, we first collect a large set of states, $\hat{\mathcal{S}}$ by letting the agent explore the environment with uniform action selection and a large episode length. We then randomly sample states from $\hat{\mathcal{S}}$ and then query the user for its label until  $\mathcal{N}^{-}$ negative examples have been collected. Once the data collection is complete, we learn a binary classifier for the concept $C_{\mathbb{X}}$. We denote the classifier learned using informed queries for the any concept $C_{\mathbb{X}}$ as $C_{\mathbb{X}}^{*}$. 

\subsection{Classifier trained with data augmentation, $\hat{C}_{\mathbb{X}}$}
\label{c_hat}
\begin{wrapfigure}{r}{0.15\textwidth}
  \begin{center}
    \includegraphics[scale=0.15]{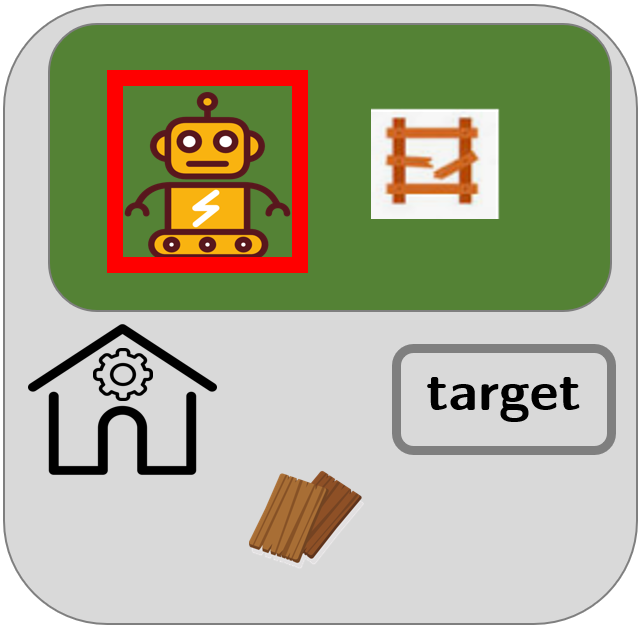}
  
\caption{Annotated patch corresponding to the concept \textit{in\_storage\_area}}
\label{fig:3}
  \end{center}
  
\end{wrapfigure}

While training a classifier with data collected using informed queries is already feedback efficient, we can further increase the efficiency using a data augmentation approach. For this approach, we assume that for any positive example state of concept $C_{\mathbb{X}}$, there is some local region within the image corresponding to the state which indicates the presence of the concept in that state. For example, in Fig \ref{fig:3}, the region within the red bounding box indicates the fact that the robot is in the storage area.  In our data augmentation technique, we ask the user to provide a minimum bounding box that covers the region that indicates the presence of the concept. Given multiple such relevant patches for the concept, we proceed to augment our dataset by sampling states from the state space and assigning them a positive label, if they have a patch that is similar to any of the patches given by the user and assign them a negative label otherwise. To operationalize this, we start with the seed examples of concept $C_{\mathbb{X}}$. From $Seed_{C_{\mathbb{X}}}$, we sample $\mathcal{W}$ states and then ask the user to annotate each sampled state with a patch that is relevant to the concept $C_{\mathbb{X}}$. Let $Patch_{C_{\mathbb{X}}}$ be the set of patches given by the user and let $S_{aug}$ be a large set of state images randomly sampled from the state space. We now require a way to measure the similarity between any two patches of the same size. Convolutional neural networks (CNNs) have been shown to be adept at finding perceptual similarity between images \cite{zhang2018unreasonable, mechrez2018contextual, amir2021understanding}. Thus, for each unique patch size $k$ among all the patches $\mathcal{P} \in Patch_{C_{\mathbb{X}}}$, we train a unique autoencoder, $\Phi_{k}$. Here $\Phi_{k}$ is trained on all possible patches of size $k$ extracted from each image in $S_{aug}$. One can reduce the number of autoencoders to be trained by scaling patches of similar size to be of same size and then train a single autoencoder for that size. Now, for any input patch $P$  let $\Phi_{k}(P)$ denote the embedding obtained as the output of the encoder in $\Phi_{k}$. Then, given two patches $P_{1}$ and $P_2$ of size $k$, we use the cosine similarity between $\Phi_{k}(P_1)$ and $\Phi_{k}(P_2)$ as the similarity measure between the patches. We now define a similarity score $Sim(P, \mathcal{I})$ between a patch ${P}$ and an image $\mathcal{I}$. Let $k_{P}$ be the size of patch ${P}$. Then, $Sim({P}, \mathcal{I})$ is given as the maximum similarity between the patch ${P}$ and all possible patches, $\mathcal{I}_{P}$,  of size $k_{P}$ extracted from $\mathcal{I}$. Formally, $Sim({P}, \mathcal{I}) = \max_{\mathcal{I}_{P}} \Phi_{k_{P}} (P) \cdot \Phi_{k_{P}} (\mathcal{I}_{P})$.

Now given some patch $\mathcal{P} \in Patch_{C_{\mathbb{X}}}$, we require images in $S_{aug}$ that has a patch similar to $\mathcal{P}$. To do so, we simply rank each image, $\mathcal{I} \in S_{aug}$ based on $Sim(\mathcal{P}, \mathcal{I})$ from most similar to least similar image. We then pick the top ${M}^{+}$  images as the ones with similar patches. We use this approach to select the top $M^{+}$ images for each patch in $Patch_{C_{\mathbb{X}}}$,  and take their union. This final set of images is treated as positive examples in the augmented data. On the other hand, for negative examples, we require images in $S_{aug}$ that has no patch that matches with patches in $Patch_{C_{\mathbb{X}}}$. To get this set, we rank each image $\mathcal{I} \in S_{aug}$ based on an aggregate similarity measure, $\overline{ Sim(\mathcal{P}, \mathcal{I})}$, which is computed as the $ Sim(\mathcal{P}, \mathcal{I})$ value averaged across all $\mathcal{P} \in Patch_{C_{\mathbb{X}}}$. We then pick the ${M}^{-}$ least similar images as the set of negative examples for the augmented set. Given \textit{concept rarity} (Section \ref{assumptions}), we keep $M^{+}$ to be a small value and $M^{-}$ to be large. Once we have the augmented dataset, we use it to train the classifier for $C_{\mathbb{X}}$. We denote the classifier learned using data augmentation for any concept $C_{\mathbb{X}}$ as $\hat{C}_{\mathbb{X}}$. Note that the labels produced by data augmentation can be noisy. Thus, $\hat{C}_{\mathbb{X}}$ could be less accurate than $C_{\mathbb{X}}^{*}$.  We handle this limitation in the approach presented in the next section.

\RestyleAlgo{ruled}
\SetKwComment{Comment}{/* }{ */}

\subsection{Data collection approach
\label{approach}}

\begin{algorithm}
\scriptsize
\caption{Algorithm to train classifier for target concept, $C_{\mathbb{T}}$}\label{alg:one}
\SetKwInOut{Input}{Input}
\SetKwInOut{Output}{Output}
\SetKw{Break}{break}
\Input{$\mathcal{M_{\mathbb{T} \rightarrow \mathbb{K}}}$, $\mathcal{P_{\mathbb{T} \rightarrow \mathbb{K}}} = \langle C_{\mathbb{T}} \rightarrow \dots C_{\mathbb{X}}  \dots \rightarrow C_{\mathbb{K}} \rangle$}
\Output{classifier  $C_{\mathbb{T}}$}
	$Seed_{C_{\mathbb{T}}} \gets $ Get\_all\_seed($C_{\mathbb{T}}$, $\mathcal{P_{\mathbb{T} \rightarrow \mathbb{K}}}$) \tcp*[l]{Algo2}
   $C_{\mathbb{T}} \gets $ Train\_classifier\_$C^{*}$($Seed_{C_{\mathbb{T}}}$) \tcp*[l]{$\S$ \ref{c_star}}
   return $C_{\mathbb{T}}$ \;
\end{algorithm}

We will now describe our data collection approach that outputs the concept classifier for the target concept $C_{\mathbb{T}}$ given the path $\mathcal{P_{\mathbb{T} \rightarrow \mathbb{K}}}$.  We first describe the simpler case when the length of the path,  $|\mathcal{P_{\mathbb{T} \rightarrow \mathbb{K}}}| = 1$. In this case, we simply collect the seed examples as described in section \ref{seed_algo}, and then proceed to learn the classifier $C_{\mathbb{T}}^{*}$ using informed queries (section \ref{c_star}). Note that we require a highly accurate classifier for $C_{\mathbb{T}}$ and thus avoid noisy data augmentation. The case of $|\mathcal{P_{\mathbb{T} \rightarrow \mathbb{K}}}| > 1$ is more involved. Let the path from the concept $C_{\mathbb{T}}$ to $C_{\mathbb{K}}$ in  $\mathcal{M}_{\mathbb{T} \rightarrow \mathbb{K}}$ be of the form $C_{\mathbb{T}} \rightarrow C_{\mathbb{I}_1} \rightarrow C_{\mathbb{I}_2} \dots C_{\mathbb{I}_n} \rightarrow C_{\mathbb{K}}$. We require some seed examples of $C_{\mathbb{T}}$ and then can use informed queries to train classifier of type $C_{\mathbb{T}}^{*}$. To do this, we need the classifier for concept $ C_{\mathbb{I}_1}$. However, in order to learn the classifier for $C_{\mathbb{I}_1}$, we need the classifier for $C_{\mathbb{I}_2}$ and so on. Thus we need to learn the concept classifiers for the intermediate concepts. Note that we only require the intermediate concept's classifiers to be accurate enough that it allows us to get some minimum number of seed examples of $C_{\mathbb{T}}$. This means that, potentially, we do not need highly accurate classifiers for the intermediate concepts. This gives us an opportunity to use the feedback efficient, albeit less accurate, $\hat{C}$ type classifier. In \textit{PRESCA}, we have developed an online approach that maintains a confidence over the ability of $\hat{C}$ classifier to obtain seed examples, and switch to training the $C^{*}$ type classifier when the confidence is low.

  \begin{algorithm}
  \scriptsize
   \caption{ Get\_all\_seed (gets required seed examples of input concept)}\label{alg:two}
   \SetKwInOut{Input}{Input}
\SetKwInOut{Output}{Output}
\SetKw{Break}{break}
\Input{$C_{\mathbb{X}}$, $\mathcal{P_{\mathbb{X} \rightarrow \mathbb{K}}}$, $\mathcal{U}$, $N^{+}$, $N^{-}$}
\Output{$Seed_{C_{\mathbb{X}}}$}
$C_{\mathbb{Y}} \gets$ Child\_concept\_of\_$C_{\mathbb{X}}$ \ $,$  $Seed_{C_{\mathbb{X}}} \gets \{\}$ \;
\eIf{$C_{\mathbb{Y}}$ is known}{
	\While{$|Seed_{C_{\mathbb{X}}}| < \mathcal{U}$}{
   		 	$s \gets $ Get\_single\_seed ($C_{\mathbb{X}}$, $C_{\mathbb{Y}}$ ) \tcp*{Algo3}
            \eIf{$C_{\mathbb{Y}}$ is a classifier $\lor$ $C_{\mathbb{X}}$ not the only cause of $C_{\mathbb{Y}}$}{
            	$C_{\mathbb{X}}(s) \gets $ Query\_user\_label ($s$) \;
            }{
            	$C_{\mathbb{X}}(s) \gets $ \textit{True}
            }
            \If{$C_{\mathbb{X}}(s)$}{
            	$Seed_{C_{\mathbb{X}}} = Seed_{C_{\mathbb{X}}} \cup s$ \;
            }
    }
}{
	\tcp*[h]{recursive call}\;
	$Seed_{C_{\mathbb{Y}}} \gets $ Get\_all\_seed($C_{\mathbb{Y}}$, $\mathcal{P_{\mathbb{Y} \rightarrow \mathbb{K}}}$)  \;
    $\hat{C}_{\mathbb{Y}} \gets$ Train\_classifier\_$\hat{C}$($Seed_{C_{\mathbb{T}}}$)   \tcp*[l]{$\S$ \ref{c_hat}}
    $\mathcal{S}_{C_{\mathbb{Y}}}^{+}$, $\mathcal{S}_{C_{\mathbb{Y}}}^{-} \gets \{\}, \{\}$ \tcp*[h]{$+$ and $-$ examples of $C_{\mathbb{Y}}$}
    \While{\textit{True}}{
    	$p \gets $ Uniform\_distribution($0$, $1$) \;
        \eIf{$p < \epsilon$}{
        	$s \gets $ Get\_single\_seed ($C_{\mathbb{X}}$, $\hat{C}_{\mathbb{Y}}$ ) \tcp*{Algo3}
            $C_{\mathbb{X}}(s) \gets $ seed\_query ($s$) \;
            \If{$C_{\mathbb{X}}(s)$}{
            	$Seed_{C_{\mathbb{X}}} = Seed_{C_{\mathbb{X}}} \cup s$ \;
            }
            Update $\epsilon$ using $\epsilon = (1-\lambda) \cdot \epsilon_{old} + \lambda \cdot \epsilon_{target}$ \;
        }{
        	$\mathcal{S}_{C_{\mathbb{Y}}}^{+}$, $\mathcal{S}_{C_{\mathbb{Y}}}^{-} \gets $ Informed\_queries () \tcp*{$\S$ \ref{c_star}}
        }
        \lIf{ $|Seed_{C_{\mathbb{X}}}| = \mathcal{U}$}{break}
        \If{$|\mathcal{S}_{C_{\mathbb{Y}}}^{+}| \geq N^{+}$ \& $|\mathcal{S}_{C_{\mathbb{Y}}}^{-}| \geq N^{-}$}{
        $C_{\mathbb{Y}} \gets$ Train\_classifier($\mathcal{S}_{C_{\mathbb{Y}}}^{+}$, $\mathcal{S}_{C_{\mathbb{Y}}}^{-}$) \;
        	$Seed_{C_{\mathbb{X}}} =$ Get\_all\_seed($C_{\mathbb{X}}$, $\mathcal{P_{\mathbb{X} \rightarrow \mathbb{Y}}}$) \; 
    	}
        }
    }
return $Seed_{C_{\mathbb{X}}}$
  \end{algorithm}

  \begin{algorithm}
\scriptsize
\caption{Get\_single\_seed}\label{alg:three}
\SetKwInOut{Input}{Input}
\SetKwInOut{Output}{Output}
\SetKw{Break}{break}
\Input{$C_{\mathbb{X}}$,$C_{\mathbb{Y}}$}
\Output{potential seed example $s$ of $C_{\mathbb{X}}$ given $C_{\mathbb{X}} \rightarrow C_{\mathbb{Y}}$ }
	\While{\textit{True}}{
    $s \gets $ Env.InitialState() \;
   	\For{$t \gets 0$ \KwTo episode\_length}{
        	 $a \gets $ UniformActionSelection() \; $s' \gets $ Env.ExecuteAction($a$)\;
            \If{$C_{\mathbb{Y}}(s') \land \neg C_{\mathbb{Y}}(s)$}{
                 return $s$ \tcp*[h]{potential example of $C_{\mathbb{X}}$}
             
            }
            $s \gets s'$
        }
    }
\end{algorithm}

The algorithm starts by getting seed examples of $C_{\mathbb{I}_n}$ which is possible because its immediate child $C_{\mathbb{K}}$ is known. Using the $Seed_{C_{\mathbb{I}_n}}$ we first train the classifier $\hat{C}_{\mathbb{I}_n}$ (section \ref{c_hat}). We now define a notion of \textit{seed} query and when such a query is \textit{successful}. Let $d$ be all the disjuncts in $\mathcal{V}_{C_{\mathbb{I}_n}}$. Say we use $\hat{C}_{\mathbb{I}_n}$ to get a potential seed example, $\mathcal{S}$, of $C_{\mathbb{I}_{n-1}}$ by finding a transition $\langle \mathcal{S}, a, \mathcal{S'} \rangle$ where $\hat{C}_{\mathbb{I}_n}(\mathcal{S})$  is \textit{false} and $\hat{C}_{\mathbb{I}_n}(\mathcal{S'})$ is \textit{true} (section \ref{seed_algo}). Then a \textit{seed} query is defined as the query where we ask the user whether any of the concept in $d$ are \textit{true} in $\mathcal{S}$. If it is indeed the case, then we say that the query was \textit{successful}. The \textit{success} rate of such queries will depend upon the accuracy of learned classifier $\hat{C}_{\mathbb{I}_n}$. Our algorithm starts by assigning some initial value to this \textit{success} rate denoted as $\epsilon$. Then with probability $\epsilon$, the algorithm will use $\hat{C}_{\mathbb{I}_n}$ to get a potential seed example of $C_{\mathbb{I}_{n-1}}$ and perform a \textit{seed} query. And with probability $1 - \epsilon$, the algorithm will perform informed queries (section \ref{c_star}) to train  $C_{\mathbb{I}_n}^{*}$. Each time a \textit{seed} query is made, the success rate is updated using a step-size $\lambda$ with  a soft target update defined as: $\epsilon = (1-\lambda) \cdot \epsilon_{old} + \lambda \cdot \epsilon_{target}$ where $\epsilon_{old}$ is the value of \textit{success} rate before the update and $\epsilon_{target}$ is the \textit{success} rate computed over all the \textit{seed} queries done so far. For every \textit{successful} query, we also query whether $C_{\mathbb{I}_{n-1}}$ is \textit{true} in the queried state. Our algorithm keeps performing these queries and terminate when either the required seed examples of $C_{\mathbb{I}_{n-1}}$ have been gathered or if $\mathcal{N}^{-}$ $\&$  $\mathcal{N}^{+}$ examples of $C_{\mathbb{I}_n}$ have been obtained (section \ref{c_star}). In the latter case, we simply train $C_{\mathbb{I}_n}^{*}$which is then used to get the set $Seed_{C_{\mathbb{I}_{n-1}}}$. Thus, if the \textit{success} rate is high, our algorithm avoids taking unnecessary feedback from the user. If that's not the case, then it switches to performing informed queries and learning more accurate classifier for the intermediate concept.  Once $Seed_{C_{\mathbb{I}_{n-1}}}$ has been collected, the algorithm will then learn the classifier $\hat{C}_{\mathbb{I}_{n-1}}$ and collect $Seed_{C_{\mathbb{I}_{n-2}}}$ using the aforementioned procedure. This process is repeated till $Seed_{C_{\mathbb{T}}}$ has been collected. Subsequently, the algorithm learns the classifier $C_{\mathbb{T}}^{*}$. We provide the complete algorithm of learning the classifier in Algorithm \ref{alg:one}.

\subsection{Using the learned concept during training}
\label{training}
Once we have the classifier for $C_{\mathbb{T}}$ we want to use it during the agent's training to incorporate the user's preference. As stated earlier, we support preference of the type in which the user wants the agent to avoid states where $C_{\mathbb{T}}$ is \textit{true}. For this case, we simply modify the MDP $\mathcal{M}$'s reward function $R$ to $R'$ such that the agent receives a high negative reward, $r_{\mathbb{T}}$ when visiting any state where  $C_{\mathbb{T}}$ is \textit{true}. We rely on $r_{\mathbb{T}}$ having a large enough negative value such that any optimal policy would achieve the goal while avoiding states with $C_{\mathbb{T}}$ \textit{true}. Our system can be easily extended to support more complex preferences. For e.g., to learn a policy in which the agent should preferably visit a state where $C_{\mathbb{T}}$ is \textit{true}, we can extend the state to have an additional \textit{flag} variable that becomes \textit{true}, once the agent visits a state with $C_{\mathbb{T}}$. We can then assign a positive reward for any transition which makes \textit{flag} \textit{true}. Once the \textit{flag} variable becomes \textit{true}, it must remain \textit{true} for every transition to avoid cycles in policy. 

\section{Evaluation}
\label{evaluation}
% \vspace{-14pt}
% \begin{wraptable}{R}{0.6\textwidth}
% \begin{table}
% \scriptsize
%   \begin{center}
%     \label{tab:table1}
%     \begin{tabular}{ |p{0.90cm}|p{0.75cm}|p{0.75cm}|p{0.90cm}|p{0.75cm}|p{0.75cm}|  } 
% \hline
% Technique & Chain length & Goal achieved & Preference aligned & Queries & patches \\
% \hline
% \multirow{3}{4em}{PRESCA} & 1 &  89\% & 100\% & 27.57 & 2k \\ 
% & 2  & 79\% & 100\% & 26.87 & 3.5k \\ 
% & 3  & 88\% & 100\% & 31.03 & 5k\\ 
% \hline
% Baseline & - & 88\% & 0\% & 28.30 & -\\
% \hline
% \end{tabular}
% \caption{Results using \textit{PRESCA} and baseline approach}
%   \end{center}
%   \vspace{-7pt}
% \end{table}

We validate \textit{PRESCA} by using it on the Minecraft domain (figure \ref{fig:2}). We have $k=10$ possible initial states for the domain in which the positions of all objects is randomly selected. These states correspond to $10$ unique maps for the domain. Recall that the goal of the agent is to drop a ladder at the target. The user's preference is that the agent must avoid going into the storage area. Thus, the target concept for \textit{PRESCA} is \textit{in\_storage\_area}.  The rewards for the original domain are set up in a way that the optimal policy will violate the user's preference. Thus, if \textit{PRESCA} is able to learn a policy that completes the task while following the user's preference, then we can correctly attribute this result to \textit{PRESCA}. We provide detailed information about the domain in appendix \ref{appendix1}. We consider the case where \textit{PRESCA} has been given the causal model shown in figure \ref{fig:2}(b). We show the performance of our approach under multiple settings wherein we vary the number of intermediate concepts that are present between the target concept \textit{in\_storage\_area} and the known concept. In each setting, we set the initially known concept as either \textit{has\_broken\_ladder}, \textit{has\_ladder} or \textit{ladder\_at\_docker} which corresponds to the number of intermediate concepts being $0, 1, $ and $2$ respectively. We use a simulated human as the user and it accurately answers queries that ask whether a concept is present in a given state. The simulated human is also used to obtain the patch indicating the presence of any concept (section \ref{c_hat}). For the concepts in Minecraft domain, this patch is taken as the minimum bounding box enclosing the agent. Some example patches have been shown in the appendix \ref{appendix1}. 

In each experimental setting, we use \textit{PRESCA} that learns the target concept \textit{in\_storage\_area} given the known concept and subsequently trains an RL agent to align with the user's preference. For the autoencoder, and concept classifiers, we train a convolutional neural network on RGB image representation of the state. For training the agent, we used tabular Q-learning. For evaluation, we run the trained agent on each map and report the percentage of times the agent achieved the goal as well as the percentage of times the agent achieved the goal while aligning with the user's preference. We provide further details on the state representation and network architectures used in appendix \ref{appendix2}. In our experiments with each chain length, we keep the required number of positive examples, $N^{+} = 100$ and negative examples $N^{-} = 200$ for informed queries (section \ref{c_star}). We keep the number of seed examples $\mathcal{U}$, as  $\mathcal{U} = 20$ for all the intermediate concepts and $\mathcal{U} = 40$ for the target concept.  The number of seed examples for the target concept was kept higher to ensure good classifier accuracy. Finally, when applying data augmentation (section \ref{c_hat}), we keep the number of patches per intermediate concept, $\mathcal{W} = 15$  and the parameters $M^{+}$ and $M^{-}$ as $\langle M^{+} = 0.05 \cdot | S_{aug}|, M^{-} = 0.75 \cdot | S_{aug}| \rangle$ where $S_{aug}$ is a large set of state images sampled from the environment by running an agent with uniform action selection. Finally, we keep the initial estimate of the \textit{success} rate, $\epsilon = 1.0$ and the step-size, $\lambda = 0.1$ (section \ref{approach}).  

\begin{table}[h]

\scriptsize
\begin{center}
    \begin{tabular}{ |p{0.90cm}|p{0.75cm}|p{0.75cm}|p{0.90cm}|p{0.75cm}|p{0.75cm}|} 
    \hline
Technique & \begin{tabular}[c]{@{}l@{}}Chain \\ length\end{tabular} & \multicolumn{1}{l|}{\begin{tabular}[c]{@{}l@{}}Goal \\ achieved\end{tabular}} & \multicolumn{1}{l|}{\begin{tabular}[c]{@{}l@{}}Preference \\ aligned\end{tabular}} & \begin{tabular}[c]{@{}l@{}}Label \\ queries\end{tabular} & \begin{tabular}[c]{@{}l@{}}Patch \\ queries\end{tabular} \\ \hline
\multirow{3}{*}{PRESCA} & 1 & \multirow{3}{*}{1.0} & \multirow{3}{*}{1.0} & 303 & - \\
 & 2 &  &  & 381 & 15 \\
 & 3 &  &  & 382 & 30 \\ \hline
\multirow{2}{*}{\begin{tabular}[c]{@{}l@{}}PRESCA\\ with $C^{*}$\end{tabular}} & 2 & \multirow{2}{*}{1.0} & \multirow{2}{*}{1.0} & 527 & - \\
 & 3 &  &  & 689 & - \\ \hline
Baseline & - & 1.0 & 0.0 & - & - \\ \hline
\end{tabular}
\end{center}
\caption{Results using \textit{PRESCA}, \textit{PRESCA} with only $C^{*}$ and baseline approach}
\label{table_main}
\end{table}

In table \ref{table_main}, we report the results for \textit{PRESCA} in each setting, along with two approaches: first is a baseline RL agent that was trained using the domain's original reward, and the second baseline is an ablated  version of our approach that only trains $C^{*}$ type classifiers for the intermediate concepts. For the second baseline, we kept $N^{+} = 40$ and $N^{-} = 80$ for intermediate concept classifiers (below which the classifiers weren't accurate enough to obtain seed examples). As we can see from the table, agents trained using \textit{PRESCA} are able to complete the task for each map, and they always aligns with the user's preference while the baseline RL agent always violates the user's preference. We also see that amount of feedback taken from the user increases by a small amount when the chain-length becomes $> 1$. Finally, we see that \textit{PRESCA} with only $C^{*}$ requires more feedback from the user for the same performance ($\sim 150$ and $ \sim 310$ additional queries for chain length $2$ and $3$ respectively). 

\section{Related work}

In this work, we develop a system that allows users to specify their preferences in symbolic terms to an RL agent that uses some high-dimensional inscrutable state representation to learn the task. Many recent \textit{neuro-symbolic} works have shown the usefulness of symbolic interfaces. In particular, works presented in (\cite{guan2022leveraging}; \cite{illanes2020symbolic}; \cite{kambhampati2022symbols}; and, \cite{yang2018peorl}) make use of symbolic knowledge to train the agent efficiently. The communication from the agent to the human has also been explored in \cite{sreedharan2020bridging}, that produces explanations for an RL agent's decisions in symbolic terms. As described in \cite{kambhampati2022symbols}, one of challenges of using a symbolic interface is that of expanding the preexisting symbolic vocabulary. In \textit{PRESCA}, we provide a mechanism that enables vocabulary expansion with minimal efforts from the users. In terms of the overarching objective of \textit{PRESCA}, which is to allow for preference specification to an RL agent, a closely related work is that of taskable RL presented in \cite{illanes2020symbolic} that allows for specification of goals in symbolic terms to the RL agent. However, that work assumes that all the relevant concepts are already available with the system. In contrast, \textit{PRESCA} relaxes this assumption, and allows learning of new concepts online by collecting positive and negative examples of the concept from the user. 

There are notable human-in-the-loop approaches that can also be used to specify and incorporate user's preference. These include the preference-based reinforcement learning approaches (\cite{christiano2017deep}, \cite{lee2021pebble}) that query the user about their preference between two trajectory segments. Then they learn a reward function that would produce the same ranking between trajectories as the user. This reward function is used to train the agent in solving the task. Another popular approach is the TAMER framework that involves the user giving ratings to actions taken by the agent (\cite{knox2009interactively}, \cite{warnell2018deep}). Then the agent learns the human's rating function and greedily chooses actions that maximize its value at each step.  These techniques are useful for tacit knowledge tasks where the user's preference cannot be stated in terms of concepts.  However, if the user's preference can be stated in symbolic terms, then using these approaches will be unnecessarily cumbersome for the user. Another limitation with these approaches, unlike \textit{PRESCA}, is that the user cannot provide feedback that is only specific to their preference and must provide feedback for the entire task.  This would arguably require much more feedback from the user, than it would for \textit{PRESCA} to learn the relevant concept. Also note that \textit{PRESCA} adds the learned concept to the symbolic interface thus amortizing the efforts of learning. When learning concepts, \textit{PRESCA} tries to reduce the feedback complexity. There are active learning works in machine learning literature \cite{settles2009active}, that also deal with reducing the number of labels needed to learn a classifier. However, active learning suffers when there is class imbalance (skew) \cite{kazerouni2020active} in the data which can often be severe in RL settings in cases where the target concepts are rare and the state space is huge. Moreover, \textit{PRESCA} can easily be complemented with active learning using simple strategies like refining candidate queries, produced by \textit{PRESCA}, using informativeness scores from active learning. 

\section{Conclusion and future work}
\label{planned_improvements}
This paper proposes the \textit{PRESCA} system through which everyday users can specify their preference to an AI agent. \textit{PRESCA} facilitates this communication by maintaining a symbolic interface made up of user understandable concepts. It also supports learning new concepts if needed. We show how \textit{PRESCA} leverages causal relationship between concepts and a novel data augmentation approach to make the learning process more feedback efficient. While we presented a competent methodology to operationalize \textit{PRESCA}, there are some improvements that can be made. First, for the case when there are multiple possible causes of the known concept, we intend to do an iterative clustering based approach which will identify the cluster corresponding to the target concept. Second, we would experiment with adding active learning methods to \textit{PRESCA} to see if that helps further reduce feedback. Finally, we would like to improve the similarity metric used in the data augmentation procedure by incorporating the transition information as well into the embedding via self-supervised learning approaches \cite{pieter2020}.

\bibliography{example_paper}

\begin{thebibliography}{24}
\providecommand{\natexlab}[1]{#1}
\providecommand{\url}[1]{\texttt{#1}}
\expandafter\ifx\csname urlstyle\endcsname\relax
  \providecommand{\doi}[1]{doi: #1}\else
  \providecommand{\doi}{doi: \begingroup \urlstyle{rm}\Url}\fi

\bibitem[Amir \& Weiss(2021)Amir and Weiss]{amir2021understanding}
Amir, D. and Weiss, Y.
\newblock Understanding and simplifying perceptual distances.
\newblock In \emph{Proceedings of the IEEE/CVF Conference on Computer Vision
  and Pattern Recognition}, pp.\  12226--12235, 2021.

\bibitem[Andreas et~al.(2017)Andreas, Klein, and Levine]{andreas2017modular}
Andreas, J., Klein, D., and Levine, S.
\newblock Modular multitask reinforcement learning with policy sketches.
\newblock In \emph{International Conference on Machine Learning}, pp.\
  166--175. PMLR, 2017.

\bibitem[Christiano et~al.(2017)Christiano, Leike, Brown, Martic, Legg, and
  Amodei]{christiano2017deep}
Christiano, P.~F., Leike, J., Brown, T., Martic, M., Legg, S., and Amodei, D.
\newblock Deep reinforcement learning from human preferences.
\newblock \emph{Advances in neural information processing systems}, 30, 2017.

\bibitem[Geffner \& Bonet(2013)Geffner and Bonet]{geffner2013concise}
Geffner, H. and Bonet, B.
\newblock A concise introduction to models and methods for automated planning.
\newblock \emph{Synthesis Lectures on Artificial Intelligence and Machine
  Learning}, 8\penalty0 (1):\penalty0 1--141, 2013.

\bibitem[Glymour et~al.(2016)Glymour, Pearl, and Jewell]{glymour2016causal}
Glymour, M., Pearl, J., and Jewell, N.~P.
\newblock \emph{Causal inference in statistics: A primer}.
\newblock John Wiley \& Sons, 2016.

\bibitem[Guan et~al.(2022)Guan, Sreedharan, and
  Kambhampati]{guan2022leveraging}
Guan, L., Sreedharan, S., and Kambhampati, S.
\newblock Leveraging approximate symbolic models for reinforcement learning via
  skill diversity.
\newblock \emph{arXiv preprint arXiv:2202.02886}, 2022.

\bibitem[Hadfield-Menell et~al.(2017)Hadfield-Menell, Milli, Abbeel, Russell,
  and Dragan]{hadfield2017inverse}
Hadfield-Menell, D., Milli, S., Abbeel, P., Russell, S.~J., and Dragan, A.
\newblock Inverse reward design.
\newblock \emph{Advances in neural information processing systems}, 30, 2017.

\bibitem[Helmert(2004)]{helmert2004planning}
Helmert, M.
\newblock A planning heuristic based on causal graph analysis.
\newblock In \emph{ICAPS}, volume~16, pp.\  161--170, 2004.

\bibitem[Icarte et~al.(2022)Icarte, Klassen, Valenzano, and
  McIlraith]{icarte2022reward}
Icarte, R.~T., Klassen, T.~Q., Valenzano, R., and McIlraith, S.~A.
\newblock Reward machines: Exploiting reward function structure in
  reinforcement learning.
\newblock \emph{Journal of Artificial Intelligence Research}, 73:\penalty0
  173--208, 2022.

\bibitem[Illanes et~al.(2020)Illanes, Yan, Icarte, and
  McIlraith]{illanes2020symbolic}
Illanes, L., Yan, X., Icarte, R.~T., and McIlraith, S.~A.
\newblock Symbolic plans as high-level instructions for reinforcement learning.
\newblock In \emph{Proceedings of the international conference on automated
  planning and scheduling}, volume~30, pp.\  540--550, 2020.

\bibitem[Kambhampati et~al.(2022)Kambhampati, Sreedharan, Verma, Zha, and
  Guan]{kambhampati2022symbols}
Kambhampati, S., Sreedharan, S., Verma, M., Zha, Y., and Guan, L.
\newblock Symbols as a lingua franca for bridging human-ai chasm for
  explainable and advisable ai systems.
\newblock In \emph{Proceedings of the AAAI Conference on Artificial
  Intelligence}, volume~36, pp.\  12262--12267, 2022.

\bibitem[Kazerouni et~al.(2020)Kazerouni, Zhao, Xie, Tata, and
  Najork]{kazerouni2020active}
Kazerouni, A., Zhao, Q., Xie, J., Tata, S., and Najork, M.
\newblock Active learning for skewed data sets.
\newblock \emph{arXiv preprint arXiv:2005.11442}, 2020.

\bibitem[Knox \& Stone(2009)Knox and Stone]{knox2009interactively}
Knox, W.~B. and Stone, P.
\newblock Interactively shaping agents via human reinforcement: The tamer
  framework.
\newblock In \emph{Proceedings of the fifth international conference on
  Knowledge capture}, pp.\  9--16, 2009.

\bibitem[Lee et~al.(2021)Lee, Smith, and Abbeel]{lee2021pebble}
Lee, K., Smith, L., and Abbeel, P.
\newblock Pebble: Feedback-efficient interactive reinforcement learning via
  relabeling experience and unsupervised pre-training.
\newblock \emph{arXiv preprint arXiv:2106.05091}, 2021.

\bibitem[Lyu et~al.(2019)Lyu, Yang, Liu, and Gustafson]{lyu2019sdrl}
Lyu, D., Yang, F., Liu, B., and Gustafson, S.
\newblock Sdrl: interpretable and data-efficient deep reinforcement learning
  leveraging symbolic planning.
\newblock In \emph{Proceedings of the AAAI Conference on Artificial
  Intelligence}, volume~33, pp.\  2970--2977, 2019.

\bibitem[Mechrez et~al.(2018)Mechrez, Talmi, and
  Zelnik-Manor]{mechrez2018contextual}
Mechrez, R., Talmi, I., and Zelnik-Manor, L.
\newblock The contextual loss for image transformation with non-aligned data.
\newblock In \emph{Proceedings of the European conference on computer vision
  (ECCV)}, pp.\  768--783, 2018.

\bibitem[Paszke et~al.(2017)Paszke, Gross, Chintala, Chanan, Yang, DeVito, Lin,
  Desmaison, Antiga, and Lerer]{paszke2017automatic}
Paszke, A., Gross, S., Chintala, S., Chanan, G., Yang, E., DeVito, Z., Lin, Z.,
  Desmaison, A., Antiga, L., and Lerer, A.
\newblock Automatic differentiation in pytorch.
\newblock 2017.

\bibitem[Settles(2009)]{settles2009active}
Settles, B.
\newblock Active learning literature survey.
\newblock 2009.

\bibitem[Sreedharan et~al.(2020)Sreedharan, Soni, Verma, Srivastava, and
  Kambhampati]{sreedharan2020bridging}
Sreedharan, S., Soni, U., Verma, M., Srivastava, S., and Kambhampati, S.
\newblock Bridging the gap: Providing post-hoc symbolic explanations for
  sequential decision-making problems with inscrutable representations.
\newblock \emph{arXiv preprint arXiv:2002.01080}, 2020.

\bibitem[Srinivas et~al.(2020)Srinivas, Laskin, and Abbeel]{pieter2020}
Srinivas, A., Laskin, M., and Abbeel, P.
\newblock Curl: Contrastive unsupervised representations for reinforcement
  learning, 2020.
\newblock URL \url{https://arxiv.org/abs/2004.04136}.

\bibitem[Sutton \& Barto(2018)Sutton and Barto]{sutton2018reinforcement}
Sutton, R.~S. and Barto, A.~G.
\newblock \emph{Reinforcement learning: An introduction}.
\newblock MIT press, 2018.

\bibitem[Warnell et~al.(2018)Warnell, Waytowich, Lawhern, and
  Stone]{warnell2018deep}
Warnell, G., Waytowich, N., Lawhern, V., and Stone, P.
\newblock Deep tamer: Interactive agent shaping in high-dimensional state
  spaces.
\newblock In \emph{Proceedings of the AAAI conference on artificial
  intelligence}, volume~32, 2018.

\bibitem[Yang et~al.(2018)Yang, Lyu, Liu, and Gustafson]{yang2018peorl}
Yang, F., Lyu, D., Liu, B., and Gustafson, S.
\newblock Peorl: Integrating symbolic planning and hierarchical reinforcement
  learning for robust decision-making.
\newblock \emph{arXiv preprint arXiv:1804.07779}, 2018.

\bibitem[Zhang et~al.(2018)Zhang, Isola, Efros, Shechtman, and
  Wang]{zhang2018unreasonable}
Zhang, R., Isola, P., Efros, A.~A., Shechtman, E., and Wang, O.
\newblock The unreasonable effectiveness of deep features as a perceptual
  metric.
\newblock In \emph{Proceedings of the IEEE conference on computer vision and
  pattern recognition}, pp.\  586--595, 2018.

\end{thebibliography}
\bibliographystyle{icml2023}

%%%%%%%%%%%%%%%%%%%%%%%%%%%%%%%%%%%%%%%%%%%%%%%%%%%%%%%%%%%%%%%%%%%%%%%%%%%%%%%
%%%%%%%%%%%%%%%%%%%%%%%%%%%%%%%%%%%%%%%%%%%%%%%%%%%%%%%%%%%%%%%%%%%%%%%%%%%%%%%
% APPENDIX
%%%%%%%%%%%%%%%%%%%%%%%%%%%%%%%%%%%%%%%%%%%%%%%%%%%%%%%%%%%%%%%%%%%%%%%%%%%%%%%
%%%%%%%%%%%%%%%%%%%%%%%%%%%%%%%%%%%%%%%%%%%%%%%%%%%%%%%%%%%%%%%%%%%%%%%%%%%%%%%
\newpage
\appendix
\onecolumn
\section{Domains details}
\label{appendix1}
We evaluated {PRESCA} on the Minecraft domain illustrated in figure \ref{fig:2} with a grid size of  $7$ X $7$ . Below we provide the details of the domain dynamics and the reward function used for the domain, as well as the modified reward function used by \textit{PRESCA}.  We also provide the state representation used for the classifier and the agent training. 

\paragraph{Domain dynamics} The goal of the Minecraft domain (figure \ref{fig:2}) is to place a ladder at the docker. There are two ways for the agent to obtain a ladder. One way is to collect planks from the environment, and craft a ladder at the crafting station. The other way is for the agent to pick up a broken ladder and repair it at the crafting station. To obtain the broken ladder, the agent must move into the storage area. Once the agent has a ladder, it must move to the cell adjacent to the docker, align itself in the direction of the docker and drop the ladder. 

\paragraph{Reward structure} The agent gets a reward of $1$ to achieve the goal. There is also a step cost of $0.0045$. To make sure that the optimal policy violates the user's preference in the default environment, we provide an additional reward of 2 if the agent picks up a broken ladder. This was done to validate that it was due to \textit{PRESCA} that the agent learns to align with the user's preference. When using \textit{PRESCA}, once the classifier for \textit{in\_storage\_area} has been learned, the agent gets a negative reward of $-2$ whenever it visits a state where \textit{in\_storage\_area} is predicted to be \textit{true}. 
 
\begin{figure}[h!]
\centering
\includegraphics[scale=0.2]{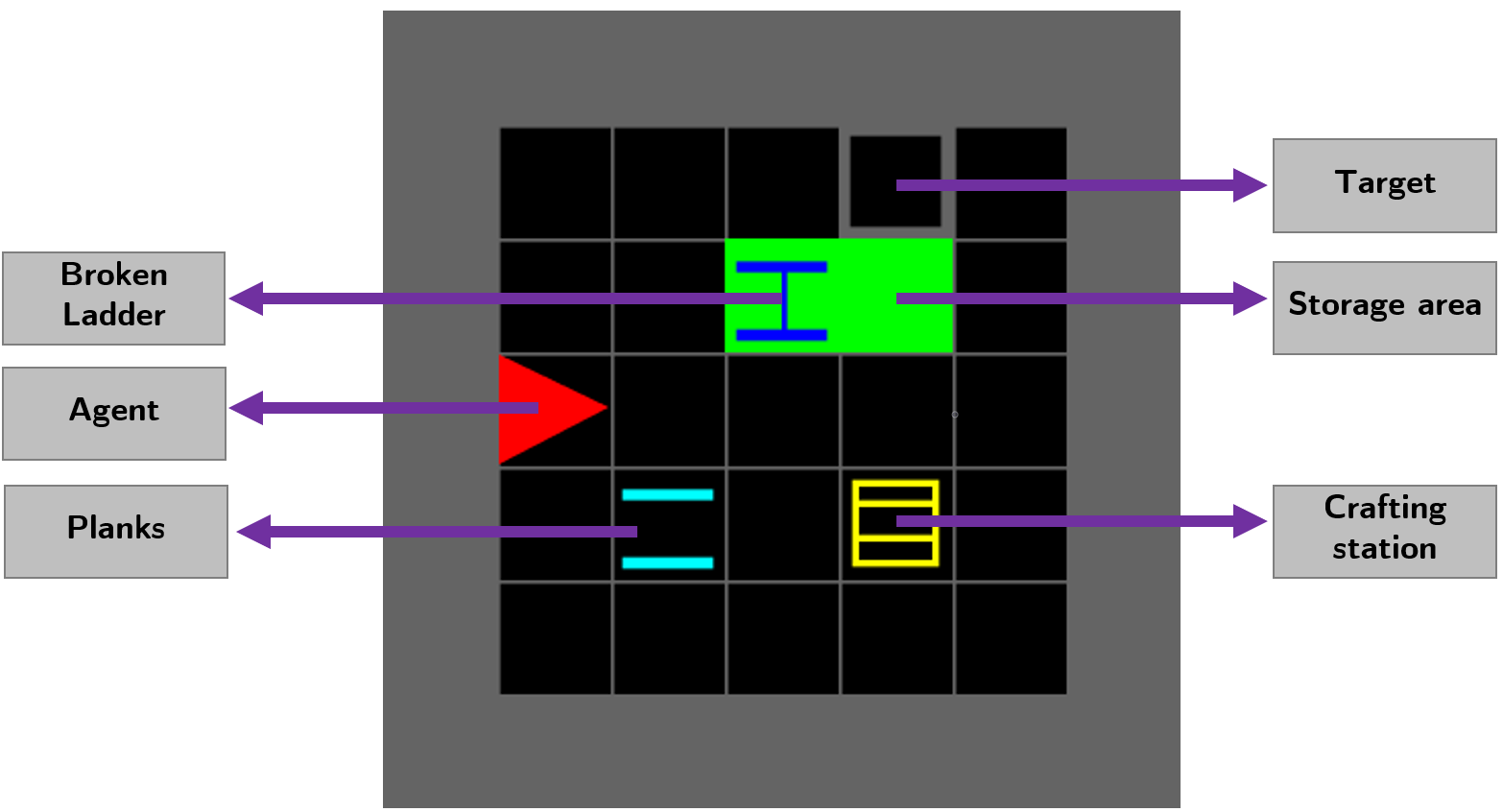}
\caption{An example state image of the Minecraft domain with all objects annotated}
\label{fig:4}
\end{figure}

\begin{figure}[h!]
\centering
\includegraphics[scale=0.2]{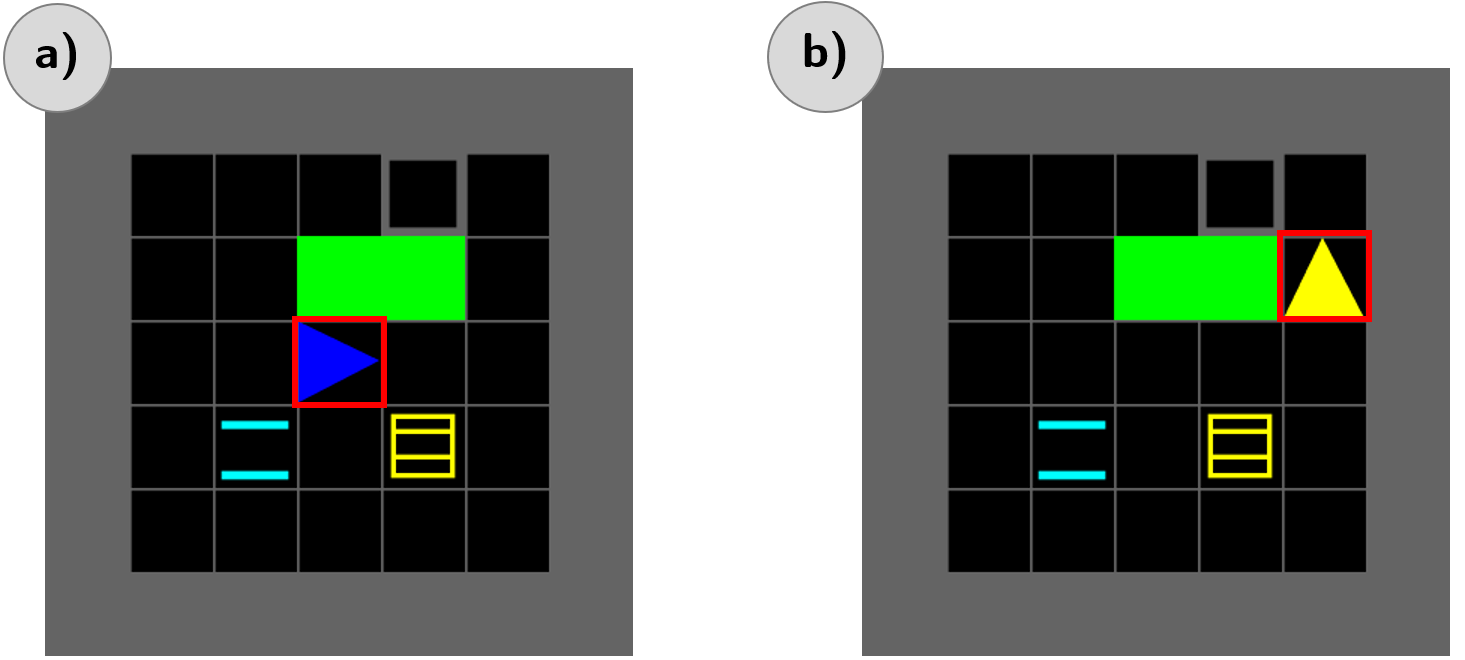}
\caption{Example patches annotated with red bounding box for the concept (a) \textit{has\_broken\_ladder} and (b) \textit{has\_ladder}}
\label{fig:5}
\end{figure}

\paragraph{Domain Images and concept patches} 
We present the domain image and annotate the objects in the image in figure \ref{fig:4}. We also provide a subset of concept patches that were used for the data augmentation algorithm (section \ref{c_hat}) in figure  \ref{fig:5}. 

\section{Training details}
\label{appendix2}

% \begin{table}[h!]
% \centering
% \small
% \caption{Network architecture for concept classifier and networks used for training the agent}
% \label{tab:table2}
% \begin{tabular}{|l|l|l|l|l}
% \cline{1-4}
%                    & Concept classifier    & Agent's policy network & Agent's actor network &  \\ \cline{1-4}
% Convolution layers & {[}Conv2D(k=2, c=16), & {[}Conv2D(k=2, c=16),  & {[}Conv2D(k=2, c=16), &  \\
%                    & MaxPool(2, 2),        & MaxPool(2, 2),         & MaxPool(2, 2),        &  \\
%                    & Conv2D(k=2, c=64)     & Conv2D(k=2, c=64){]}   & Conv2D(k=2, c=64){]}  &  \\
%                    & , MaxPool(2, 2){]}    &                        &                       &  \\ \cline{1-4}
% Linear layers      & {[}Linear(h=16)       & {[}Linear(h=64)        & {[}Linear(h=64)       &  \\
%                    & , Linear(o=2){]}      & , Linear(o=1){]}       & , Linear(o=7){]}      &  \\ \cline{1-4}
% \end{tabular}
% \end{table}

\begin{table}[h!]
\centering
\small

\begin{tabular}{|l|l|}
\cline{1-2}
                    Concept classifier    & Autoencoder network     \\ \cline{1-2}
                 {[}Conv2D(k=3, c=16), & {[}Conv2D(k=4, c=8),    \\
                 BatchNorm(c=16),        & BatchNorm(c=8),            \\
                 MaxPool(2, 2),     & Conv2D(k=4, c=16),      \\
                 Conv2D(k=3, c=32),    & BatchNorm(c=16),       \\ 
                  BatchNorm(c=32),                  & Conv2D(k=4, c=16),      \\ 
                 MaxPool(2, 2),                     & BatchNorm(c=16),       \\ 
                 Conv2D(k=3, c=64),                    & ConvTranspose2d(k=3, c=16),      \\ 
                 BatchNorm(c=64),                          & BatchNorm(c=16),      \\ 
                 MaxPool(2, 2),                          & ConvTranspose2d(k=3, c=8),       \\ 
                 Linear(d=256),                          & BatchNorm(c=8),       \\ 
                 Sigmoid(Linear(d=1)){]}                          & ConvTranspose2d(k=4, c=3){]}     \\ \cline{1-2}
\end{tabular}
\caption{Network architecture for concept classifier and the autoencoder used}
\label{table2}
\end{table}

The concept classifiers were trained on RGB image representation of the state that is of shape $56 \text{ X } 56 \text{ X } 3$. For training the concept classifier, we use a convolutional neural network with $20$ epochs of training over the data. For training, we used stochastic gradient descent with a learning rate of $0.001$. The training is done to minimize the cross-entropy loss. For similarity score calculation during data augmentation (section \ref{c_hat}), we trained the autoencoders on patches of size $k=8 \text{ X } 8$ with 3 channels. The patches are extracted from input image $\mathcal{I} \in S_{aug}$ with a stride of 1 where $S_{aug}$ is a large set of state images sampled from the environment. We used Mean-squared error (MSE) between the input patch and reconstructed output as the network loss. The autoencoder is trained for $20$ epochs using Adam optimizer and learning rate of $0.001$. All the models were implemented with PyTorch \cite{paszke2017automatic}. The network architecture used for concept classifiers and the autoencoder is provided in table \ref{table2}.
%%%%%%%%%%%%%%%%%%%%%%%%%%%%%%%%%%%%%%%%%%%%%%%%%%%%%%%%%%%%%%%%%%%%%%%%%%%%%%%
%%%%%%%%%%%%%%%%%%%%%%%%%%%%%%%%%%%%%%%%%%%%%%%%%%%%%%%%%%%%%%%%%%%%%%%%%%%%%%%

\end{document}